\pdfoutput=1
\documentclass[11pt]{article}
\usepackage{booktabs}
\usepackage{float}
\usepackage[dvipsnames]{xcolor}
\usepackage[many]{tcolorbox}
\usepackage[everyline=true,framemethod=tikz]{mdframed}

\usepackage[final]{acl}
\usepackage{etoolbox}
\gappto{\UrlBreaks}{\UrlOrds}
\usepackage{times}
\usepackage{arydshln}
\usepackage{latexsym}
\usepackage{amsmath}
\usepackage{multirow}
\usepackage{enumitem}
\usepackage[symbol]{footmisc}
\usepackage[T1]{fontenc}

\usepackage[utf8]{inputenc}

\usepackage{microtype}

\usepackage{inconsolata}

\usepackage{graphicx}
\usepackage{amssymb}
\usepackage{pifont}
\newcommand{\cmark}{\ding{51}}%
\newcommand{\xmark}{\ding{55}}%
\title{Seemingly Plausible Distractors in Multi-Hop Reasoning:\\Are Large Language Models Attentive Readers?}

\newcommand*{\affaddr}[1]{\normalfont{#1}} 
\newcommand*{\affmark}[1][*]{\textsuperscript{#1}}
\newcommand*{\email}[1]{\texttt{#1}}
\author{
Neeladri Bhuiya\affmark[2] \And
Viktor Schlegel\affmark[3,4]\footnotemark[1] \And
Stefan Winkler\affmark[1,2]
\AND 
\affaddr{\affmark[1]ASUS Intelligent Cloud Services (AICS) Singapore}, 
\affaddr{\affmark[2]National University of Singapore} \\
\affaddr{\affmark[3]University of Manchester, United Kingdom},
\affaddr{\affmark[4]Imperial Global Singapore} \\
\email{\small{neeladri.bhuiya@u.nus.edu}}, \email{\small{v.schlegel@imperial.ac.uk}},
\email{\small{winkler@nus.edu.sg}}\\
}

\begin{document}
\maketitle
\footnotetext[1]{Work done while author was at AICS.}
\begin{abstract}
State-of-the-art Large Language Models (LLMs) are accredited with an increasing number of different capabilities, ranging from reading comprehension over advanced mathematical and reasoning skills to possessing scientific knowledge. In this paper we focus on multi-hop reasoning---the ability to identify and integrate information from multiple textual sources. Given the concerns with the presence of simplifying cues in existing multi-hop reasoning benchmarks, which allow models to circumvent the reasoning requirement, we set out to investigate whether LLMs are prone to exploiting such simplifying cues. We find evidence that they indeed circumvent the requirement to perform multi-hop reasoning, but they do so in more subtle ways than what was reported about their fine-tuned pre-trained language model (PLM) predecessors. We propose a challenging multi-hop reasoning benchmark by generating seemingly plausible multi-hop reasoning chains that ultimately lead to incorrect answers. We evaluate multiple open and proprietary state-of-the-art LLMs and show that their multi-hop reasoning performance is affected, as indicated by up to 45\% relative decrease in F1 score when presented with such seemingly plausible alternatives. We also find that---while LLMs tend to ignore misleading lexical cues---misleading reasoning paths indeed present a significant challenge. The code and data are made available at \url{https://github.com/zawedcvg/Are-Large-Language-Models-Attentive-Readers}.
\end{abstract}

\section{Introduction}
Recent developments in the field of language modelling and the introduction of open~\cite{touvron2023llama} and proprietary~\cite{OpenAI2023GPT-4Report} Large Language Models (LLMs) have undeniably advanced the state of the art in Natural Language Processing (NLP). LLMs have been credited with various understanding and reasoning capabilities, ranging from  arithmetic~\cite{Cobbe2021TrainingProblems}, deductive~\cite{Saparov2023TestingExamples} and formal~\cite{Schlegel2022CanLanguage,Madusanka2023NotQuantifiers} reasoning and 
possessing general~\cite{AlKhamissi2022ABases}, 
and domain-specific~\cite{He2023MedEval:Evaluation} knowledge. Due to their size and generalisation capabilities~\cite{Brown2020}, their evaluation on benchmarks requiring such types of reasoning is typically performed in zero- or few-shot settings on many NLP tasks, without the need for fine-tuning datasets. 

\begin{figure}[!t]
    \centering
    \resizebox{0.95\columnwidth}{!}{%
    \begin{tabular}{|p{1.2\columnwidth}|}
    \hline
    \textbf{Original Q:} Who created the 2003 remake of the 1983 overhead view, vehicular combat game developed by Bally Midway?\\
    \hdashline
    \textbf{HotpotQA Paragraph 1:} Highway Pursuit is a computer game remake of Spy Hunter created by {\color{YellowGreen}\textbf{Adam Dawes}} [\ldots]. \\
    \textbf{HotpotQA Paragraph 2:} {\color{YellowGreen}\textbf{Spy Hunter}} is an overhead view, vehicular combat game developed by Bally Midway and released in arcades in 1983.
  \\ 
    \textbf{GPT-4 Answer}: Adam Dawes {\color{green}\cmark}\\
    \hdashline
    \textbf{Fake paragraph 1:}
     {\color{red}\textbf{Road Blaster}}, developed by Atari Corporation in 1983, stands out as a seminal entry in the vehicular combat game genre [\ldots].\\
    \textbf{Fake paragraph 2:}
    The 2003 remake of Road Blaster was masterfully recreated by {\color{red}\textbf{Jonathan Fields}} [\ldots]. \\
    \textbf{GPT-4 Answer}: Jonathan Fields {\color{red}\xmark} \\
    \hline
    \end{tabular}
    }
     \caption{Our proposed method evaluates the multi-hop reasoning capabilities of Large Language Models by adding seemingly plausible, yet ultimately wrong alternate reasoning paths, impacting the reasoning performance of state-of-the-art LLMs such as GPT-4.}
     \label{fig:introduction}
 \end{figure}

These zero- and few-shot capabilities seem to alleviate one of the weaknesses identified with the previous generation of fine-tuning based NLP architectures such as transformer-based~\cite{Vaswani2017}, and pre-trained language models~\cite{Devlin2018}---the reliance on data-set specific ``artefacts''~\cite{gururangan2018annotation,Schlegel2020a} and, as a consequence, lack of generalisation beyond specific datasets. For example, in one of the popular reading comprehension and reasoning benchmarks~\cite{Dua2019}, the majority of questions starting with ``How many'' can be answered correctly with ``2''. Following standard fine-tuning practice and splitting data in train and test randomly, such a simple heuristic will be present in both training and evaluation data, so a fine-tuned model will learn it and obtain high scores, without necessarily performing reasoning. 
LLMs seemingly circumvent this issue, as they are not fine-tuned on benchmark data. As such, they are not exposed to simplifying dataset artefacts by design, and it is reasonable to assume that they do not learn to exploit them.

However, while there is a growing body of work investigating the strengths and limitations of LLMs~\cite{Huang2023AQuestions}, little research has been carried out to validate this assumption, and to investigate whether and to what extent LLMs inherit the ``dataset artefact'' weaknesses of their fine-tuned predecessors. This is an important research question to pursue, motivated by recent findings on benchmark leakage into pre-training or instruction-tuning data~\cite{deng2024benchmark}, which invalidate the zero-shot setting and potentially allow LLMs to learn such dataset artefacts. Another line of research suggests that LLMs tend to ``over-reason'' 
\cite{Chiang2024Over-ReasoningModels}, perhaps due to ``sycophancy'' 
\cite{Perez2023DiscoveringEvaluations}, i.e., the tendency to generate the presumably preferred answer over the correct one, leading to complicated reasoning where none is required.

In this paper, we turn our attention to the well-studied capability to perform multi-hop reasoning and reading comprehension---that is, to integrate textual information from multiple different source documents. Typically, this capability is evaluated by asking questions where the necessary information to arrive at the correct answer is spread across multiple documents~\cite{Yang2018,Welbl2018,Inoue2020R4C:Reason}. It is important to understand to what extent NLP methods possess this capability, as it is required for many real-world tasks, such as retrieval-augmented generation~\cite{lewis2020retrieval} when summarising retrieved documents, and because it is a necessary prerequisite to human-level reading comprehension~\cite{Kintsch1988}.

Previous work has shown that NLP architectures might possess inadequate capabilities to perform multi-hop reasoning~\cite{min2019compositional}. However, these findings were established before the advent of large language models. To have a clear understanding of the limitations of the capabilities of state-of-the-art research, it is crucial to re-investigate these claims with the current generation of LLM-based approaches~\cite{Bowman2022TheFail}. While there is vivid research on (open-book) multi-hop reasoning capabilities of LLMs~\cite{Sakarvadia2023MemoryModels,Liu2023ArgumentTask,Yang2024DoReasoning}, how well they perform when presented with multiple, seemingly plausible multi-hop reasoning paths remains unclear.

To address this gap, we focus on the capability of LLMs to perform multi-hop reasoning when multiple seemingly plausible answers are present, where only minor details invalidate the alternative. We show that existing methods---calibrated to evaluate pre-LLM architectures---are inadequate to evaluate LLMs, and that LLM reasoning failures are indeed distinct from their fine-tuned PLM predecessors. We present a methodology to generate challenging examples with ``plausible distractors'' to evaluate LLMs' capabilities to perform multi-hop reasoning when presented with seemingly correct, but ultimately wrong and thus distracting evidence. Our results show that the reasoning capabilities of a range of open and proprietary LLMs, including GPT-4, are affected by these ``plausible distractors''. 










\section{Related Work}

It has been shown that basic pattern matching~\cite{Schlegel2020} and one-hop~\cite{min2019compositional} models can solve a large proportion of questions in multi-hop question answering datasets, presumably because the answer sentence often contains keywords common with the question, thus negating the need to follow a reasoning path and attend to multiple documents. Particularly HotpotQA~\cite{yang2018hotpotqa}, due to its multi-hop question design, was the subject of multiple studies. Approaches architecturally incapable of multi-hop reasoning still achieved close to state-of-the-art performance~\citep{min2019compositional,trivedi-etal-2020-multihop}, suggesting questions answerable in such a way do not necessitate multi-hop reasoning. 

In light of these results, several adversarial attacks have been proposed to check whether the dataset evaluates multi-hop reasoning without exhibiting ``shortcuts'', by ensuring that the correct answer can only be procured if the evaluated model can retrieve and combine information from distinct reasoning hops. 
\citet{jiang2019avoiding} elicited distracting paragraphs by using the titles of the gold paragraphs and the answer, which are subjected to phrase-level perturbations and word replacement, thus creating a distracting paragraph. Others decomposed the multi-hop questions in multiple single questions~\cite{min2019multi,perez2020unsupervised,ding2021reasoning} (e.g. DecompRC in  Figure~\ref{fig:decomp-question}) 
showed that the---typically BERT- or other PLM-based---fine-tuned SOTA models struggled to answer both sub-questions correctly when answering the complete question, or were distracted by their alterations, suggesting the presence of reasoning shortcuts~\cite{tang-ng-tung-2021-domultihop}.

By design, these methods bear only \emph{negative predictive power}~\cite{Gardner2020}: failing to see a performance drop does not imply that the model performs the evaluated capability well, but rather that the methodology might have limited suitability to evaluate the investigated phenomenon, i.e., multi-hop reasoning. As the methodologies presented above focus on fine-tuned models, 
they assume that multi-hop reasoning is circumvented through simple, lexical similarity-based methods like word matching. For example,~\citet{jiang2019avoiding} do not consider that their generated paragraphs are isolated, as they contain no explicit reference to other paragraphs in the context, such as a shared named entity. 
Meanwhile,~\citet{ding2021reasoning}\footnote{No public code/dataset was made available} 
only add a single distracting sentence. 
Thus, simple word matching, which ensures that the final answer is of the same entity type as in the question, can often lead to the correct answer. This might not be sufficient for LLMs, as they---due to their size and emergent capabilities---might circumvent multi-hop reasoning by exploiting more subtle textual cues.
Indeed, in our empirical study, we show that existing methods, due to these limitations, do not adequately test an LLM's reasoning capabilities.

Therefore, to analyse an LLM's ability to reason more adequately, we go beyond the state of the art and introduce a novel method to more effectively evaluate the multi-hop reasoning capabilities of LLMs. 
Specifically, we ensure the continuity of seemingly plausible alternative reasoning paths, which lead to answers that are ultimately wrong. To succeed, the model is required to pay close attention to small yet important details in the questions and paragraphs.

This ability is important practically, for example when an LLM is prompted to evaluate/summarise the outcome of a debate, where both sides will present plausible arguments with only one being ultimately correct~\cite{sun2023query,li2024side}. With LLMs increasingly used to judge and improve (other) LLMs' potentially similar outputs on the same topic~\cite{huang-etal-2023-large}, it is important to establish, if they possess the necessary prerequisites to do so. More broadly, similar to other works in this line of research, we look at linguistic \emph{competence} rather than \emph{performance}~\cite{chomsky1965aspects}: if we accredit multi-hop reasoning capabilities to LLMs, then, similar to humans, we expect them to exhibit these capacities not only in the majority of cases but in edge case scenarios as well, such as when presented with seemingly plausible alternate reasoning paths.

\section{Methodology}
\label{sec:methodology}
In this section, we describe our approach to evaluating the multi-hop reasoning capabilities of LLMs. We do so by creating ``distractor'' paragraphs that present seemingly plausible yet incorrect alternative paths in the reasoning chain while ensuring that this process doesn't affect the final solution. First, the question is treated as a two-hop question and converted into two sub-questions. This is done to be able to branch out alternative reasoning paths from each of the sub-questions. The sub-questions are analyzed to identify modifiable portions, 
which are then manipulated to create ``distractor'' sub-questions that lead to a different answer and thus a different reasoning chain, which is ultimately wrong, as the models are presented with the original, unmodified question.
The ``distractor sub-questions'' are finally used to generate ``distractor paragraphs'' containing ``distractor answers'' utilizing an LLM. 

The method comprises three main steps: I.~Acquiring the main entity, II.~Extracting its modifiable details, and III.~Creating the distractor paragraphs.

\begin{figure}[htb]
    \centering
    \resizebox{0.95\columnwidth}{!}{%
    \begin{tabular}{|p{1.2\columnwidth}|}
    \hline
    \textbf{Question:} What year did Guns N Roses perform a promo for a movie starring Arnold Schwarzenegger as a former New York Detective?\\
    \textbf{Sub question 1:} Which movie stars Arnold Schwarzenegger as a former New York Police detective?\\
    \textbf{Sub question 2:} What year did Guns N Roses perform a promo for \underline{\smash{End of Days}} \textit{(answer of the previous question)}? \\
    \hline
    \end{tabular}
    }
    \caption{Example of a decomposed multi-hop question.}
    \label{fig:decomp-question}
\end{figure}

\paragraph{I. Acquiring the main entity}
We use the human-annotated sub-questions from \citet{tang-ng-tung-2021-domultihop}, as exemplified in Figure~\ref{fig:decomp-question}. 
We define main entities as those that are the focus of the question. For example, in Figure~\ref{fig:decomp-question}, the main entities for the sub-questions would be ``movie stars'' and ``year'' respectively. We choose the ``main entity'' in each sub-question, using a few dependency parse-based rules\footnote{By \emph{``Dependency of type C between A and B''} we mean A is the head, B is the dependent and is C the relation type. See Appendix for type definitions.}. Intuitively, we exploit the relations between the ``wh''-word and other noun phrases to extract the main entity. Specifically: 

\begin{enumerate}[noitemsep, labelwidth=!, labelindent=0pt, topsep=3pt,label=(\roman*),font=\itshape]
    \item If the ``wh'' question word \textit{WH} is the root, and there exists a word A with a dependency \textit{nsubj} or \textit{nsubj:pass} with \textit{WH} as the head, A is the main entity.
    \item Alternatively, if there exists a word A with a dependency of type \textit{det}, \textit{nsubj}, or \textit{nsubj:pass} with a wh-word \textit{WH}:
    \begin{enumerate}[noitemsep, labelwidth=!, labelindent=0pt, topsep=0pt,font=\itshape]
        \item If A is a noun, A is the main entity.
        \item Otherwise, if A is a verb, the word B having a relation \textit{acl:recl} with B being the head, we mark B as the main entity.
    \end{enumerate}
    \item Else, if any word A has a dependency with a word B of type \textit{nsubj} or \textit{nsubj:pass}, where B is the word with a direct dependency with the wh-word, A is assigned as the main object.
\end{enumerate}
    



\paragraph{II. Extracting the details}
Next, we extract the details that need to be manipulated to create the distractor question. The main idea is to obtain modifiers of any entity in the question other than the main entity (from the previous step). Specifically:

\begin{enumerate}[noitemsep, labelwidth=!, labelindent=0pt, topsep=3pt,label=(\roman*),font=\itshape]
    \item For any dependency between two words C and D, we check if the dependency is of the form \textit{obl}, \textit{obj}, \textit{nsubj}, or \textit{nsubj:pass}. We also ensure that D isn't the main entity identified in the previous step.
    \item If the above rule is satisfied, we check if C or D has a dependency \textit{appos} with any named entity.
    \item If there is no such relation, modifiers of D of the form \textit{nummod}, \textit{amod}, \textit{nmod}, \textit{compound}, or \textit{flat} are used to get modifiable parts if the modifier isn't the main entity identified in the previous steps.
\end{enumerate}


We extract the modifiers and not the object they modify for two reasons: First, changing the object often causes the overall question to become nonsensical. Secondly, changing the modifier ensures a minimal yet semantically meaningful modification of the question \cite{Schlegel2021}. 

 \begin{figure}[ht]
    \centering
    \resizebox{\columnwidth}{!}{%
    \begin{tabular}{|p{1.2\columnwidth}|}
    \hline
    \textbf{Original Q:} The {\color{blue}\textbf{arena}} where the Lewiston Maineiacs played their {\color{red}\textbf{home}} games can seat how many people?\\
    \textbf{Sub-Q 1:} Which {\color{blue}\textbf{arena}} the Lewiston Maineiacs played their {\color{red}\textbf{home}} games?\\
    \textbf{Sub-Q 2:} How many people can the Androscoggin Bank Colisée seat? \\
    \hline
    \textbf{Fake paragraph 1:}
    The Lewiston Maineiacs took to the ice at the {\color{violet}\textbf{Maple Leaf Arena}} for their thrilling {\color{orange}\textbf{playoff}} games. [\ldots]\\
    \textbf{Fake paragraph 2:}
    {\color{violet}\textbf{Maple Leaf Arena}}, known for its state-of-the-art facilities and spacious seating, can accommodate an impressive number of {\color{cyan}{\textbf{4,500 spectators}}}. [\ldots] \\ 
    \hdashline
    \textbf{Gold Paragraph 1:} {\color{violet}\textbf{The Androscoggin Bank Colisée}} [\ldots] is a {\color{cyan}\textbf{4,000-capacity (3,677 seated)}} multi-purpose arena, in Lewiston, Maine, that opened in 1958. [\ldots] \\
    \textbf{Gold Paragraph 2:}
        The Lewiston Maineiacs [\ldots] played its {\color{red}\textbf{home}} games at the {\color{violet}\textbf{Androscoggin Bank Colisée}}. [\ldots] \\ 
    \hline
    \end{tabular}
    }
     \caption{Instantiation of our proposed method. With ``{\color{blue}\textbf{arena}}'' as main entity of sub-question 1, we extract ``{\color{red}\textbf{home}}'' to be replaced with ``{\color{orange}\textbf{playoff}}''. Then, we use the modified sequence with the original sub-question 2 (masking the answer ``Androscoggin Bank Colisée'') as prompt to \texttt{GPT-4} to generate the distractor paragraphs 1 and 2. The distractor paragraphs generated have ``{\color{violet}\textbf{Maple Leaf Arena}}'' as the bridging entity in the false reasoning chain which leads to the wrong answer ``{\color{cyan}\textbf{4500 spectators}}''.} 
     \label{fig:example-method}
 \end{figure}
 
\paragraph{III. Creating the distractor paragraphs}
After obtaining modifiable parts, we distinguish whether these are  Named Entities or not. For each of the named entities, we obtain their type  using \citet{qi2020stanza}'s Named Entity Recognition (NER) processor.  We then generate a fake entity of the same type with the help of \texttt{GPT-4}. 

Next, for the non-named entities, we use \texttt{RoBERTa's} \cite{liu2019roberta} masked token prediction objective to obtain alternative words. Specifically, we mask the modifiable parts and sample the top ten probable tokens from the language model. To ensure that the new word is sufficiently different yet still plausible given the context, we establish the following constraints empirically:
\begin{itemize}[noitemsep,wide,labelindent=0pt,topsep=5pt]
\item \textbf{Sentence Similarity} of the new sequence in comparison to the initial question, as given by the cosine similarity of \texttt{all-mpnet-base-v2} \cite{Reimers2019Sentence-BERT:BERT-Networks} is $< 0.991$;
\item \textbf{Word similarity} under \texttt{RoBERTa} of the original word and the word replacing it is $< 0.4$;
\item \textbf{Perplexity}, i.e.\ the \texttt{RoBERTa} predicted probability of the new sentence, is $> 0.001$. 
\end{itemize}

The new words and named entities are used to create new fake questions. We use these fake questions to create fake question tuples, i.e., fake questions for the different hops. While generating the fake question tuples, we mask the tokens in the second sub-question corresponding to the first sub-question's answer. Next, we feed these fake tuples into \texttt{GPT-4} and ask it to generate the distractor paragraphs. We generate a pair of distractor paragraphs for each tuple. Figure \ref{fig:example-method} shows the instantiation of our proposed method on a single example, with the generated distractor paragraphs and the corresponding gold paragraphs. In the attack each of these distractor paragraphs replaces one of the non-gold paragraphs, to prevent adding extra tokens and to ensure that the ratio of 2 gold paragraphs and 8 distractor paragraphs of the distractor setting of HotpotQA is maintained. 

\paragraph{Data Quality}
Following this procedure, we generate 132 instances of the ``other'' type, while 547 are created from named entities. To ensure that the generated distractor paragraphs are valid, do not contradict the gold paragraphs, and do not cause contradictions with the label, we randomly sample and inspect 100 named entity-based and all 132 of the ``other'' examples. For the former, none of the sampled examples were contradictory. For the latter, 13 were found to have either one or both of the distractor paragraphs contradictory---those examples were discarded. Furthermore, we conducted a user study (see Appendix~\ref{subsec:user_study}), which showed that humans have no difficulty extracting the correct answer when given a combination of real and distractor paragraphs. It was also reported that the distractor paragraphs seldom contain contradicting information. We further compare the word count of the adversarial and the original paragraphs to check if the adversarial paragraphs artificially increase complexity through a larger word count. On average, the adversarial paragraphs had a word count of 81.2, slightly lower than the average word count of the original paragraphs, which is 95.95.

Through manual verification, a user study, and the comparison of the word count of plausible paragraphs and their counterpart real paragraphs, we can conclude with high certainty that the plausible paragraphs don't contain contradictory information, and that the drop in performance of the models is due to their inherent weakness and not some artificially added complexity.

\section{Experiment Setup}
First, we investigate LLM's capabilities and limitations compared to previous PLM-based state of the art. Then, we evaluate the multi-hop reasoning capabilities of LLMs using our proposed methodology. Finally, we conduct an in-depth analysis of what makes reasoning hard for LLMs on our benchmark and conclude by evaluating state-of-the-art LLMs and prompting techniques. Unless mentioned otherwise, we use the chat models for Llama-2.

\paragraph{Do LLMs suffer from the same flaws as fine-tuned models?}
\texttt{Llama-2-13B}~\cite{touvron2023llama} is used as the baseline LLM. 
We evaluate using few-shot prompts, as these allow the model to stick to the expected output format better than zero-shot. This setting is used throughout the paper unless mentioned otherwise. Two styles of prompts were used, normal and chain of thought, as per the strategies discussed in \citet{wei2023chainofthought}. 
All reported metrics are measured at token level and averaged across all the instances, following standard evaluation practice~\citep{yang2018hotpotqa}.

We test the LLMs' performance when attacked with AddDoc~\cite{jiang2019avoiding}, an adversarial attack on HotpotQA for BERT-based models. This is intended to check an LLM's ability to handle ``distracting'' paragraphs. 
SubQA was used to determine if the models could answer the individual questions before answering the entire question. It is a sample of 1000 questions and their sub-questions from the dev set of HotpotQA, with the sub-questions being human-verified. This allows us to evaluate model \emph{consistency} in answering both the multi-hop question as well as the individual sub-questions correctly. It also allows us to investigate the opposite: When the (more complex) composite question is answered correctly, but either of the (simpler) decomposed questions is answered wrongly, the model might rely on some reasoning shortcuts, discarding sub-question information.
Finally, we evaluate if LLMs can retrieve the correct answer when necessary information from one of the gold paragraphs is missing, using the DiRe test set~\cite{trivedi-etal-2020-multihop}.

\paragraph{Do LLMs get distracted by seemingly plausible alternate reasoning paths?}
As described in Section~\ref{sec:methodology}, the attack aims to create paragraphs that provide irrelevant information that is closely related to the property/entity being questioned about. Here, we evaluate a representative sample of open-source and proprietary LLMs, specifically, \texttt{Llama-2-13B}, \texttt{Llama-2-70B}, \texttt{Mixtral-8x7B-Instruct-v0.1}, \texttt{GPT-3.5} and \texttt{GPT-4}. To contextualise the performance of LLMs to their fine-tuned PLM counterparts, we also fine-tune a \texttt{longformer} model on the HotpotQA training set and evaluate it on our proposed benchmark (see Appendix for details). Based on the chatbot leaderboard \cite{chiang2024chatbot} at the time of writing, the best state-of-the-art model was \texttt{GPT-4}. Thus we evaluate \texttt{GPT-4} to investigate how our findings generalise to stronger models.

\paragraph{What are the effects of the different parameters?}
Experiments are conducted to check the impact of the method's parameters on the performance of LLMs. Specifically, the different parameters we investigate are: 1) number of ``distractor'' paragraphs generated, i.e., two or four; 2) whether the distractor paragraphs are generated from the two sub-questions belonging to the same multi-hop question or if the sub-questions belong to two independent multi-hop questions; 3) The type of modifiable portion that is changed in the sub-question, i.e., Named Entity or not; 4) whether the paragraphs, if not generated from two distinct sub-questions, are both generated from the second sub-question.



\section{Experiment Results}
In this section, we present the results of our experiments, compare them against prior work, and discuss deeper insights. Unless otherwise stated, all reported results of the adversarial attack are statistically significant at $p<0.05$, determined by conducting a one-sided Student's t-test.

\subsection{Do LLMs suffer from the same flaws as fine-tuned models?}
\paragraph{I. Setting up the baseline}
\texttt{Llama-2-13B} chat model is used as the baseline for the performance of an LLM in a zero/few-shot setting; results are shown in Table~\ref{cot_and_control}. The F1 score indicates that the few-shot setting without chain-of-thought prompting performs best. This is because in the chain of thought setting the model often gives a lengthy explanation, thus reducing precision and F1 score.

\begin{table}[ht]
\resizebox{1\columnwidth}{!}{%
  \begin{tabular}{ccccc}
    {\bfseries Type}&{\bfseries F1 score}&{\bfseries Precision}&{\bfseries Recall}& {\bfseries Ans \# words (avg)}\\
    \hline
    Normal & 0.5077&0.5180&0.575&4.6\\
    CoT &0.4791&0.4682&0.599&5.08\\
  \hline
  \hline
\end{tabular}
}
\caption{Comparing normal and chain-of-thought prompts using \texttt{Llama-2-13B} as baseline.}
\label{cot_and_control}
\end{table}

\paragraph{II. Reasoning shortcuts using SubQA}
Table~\ref{result_SubQA} shows the result of running few-shot \texttt{Llama-2-13B} in the controlled setting on the SubQA dataset. LLama-2 performs much better on the individual sub-questions than the question requiring multi-hops. This finding, in line with analyses focusing on fine-tuned models \cite{tang-ng-tung-2021-domultihop}, suggests some inconsistencies in its reasoning capabilities and difficulty in combining information from multiple sources.  

\begin{table}[ht]
\begin{center}
\resizebox{1\columnwidth}{!}{%
  \begin{tabular}{lccc}
    \bfseries Type& \bfseries F1 score&\bfseries Precision&\bfseries Recall\\
    \hline
    Original &0.427&0.427&0.507\\
    Sub question 1&0.743&0.761&0.789\\
    Sub question 2&0.693&0.691&0.782\\
  \hline
  \hline
\end{tabular}
}
\caption{Results of \texttt{Llama-2-13B} on SubQA dataset}
\label{result_SubQA}
\end{center}
\end{table}

Table~\ref{TFF_thing} indicates the performance statistics for individual samples. $F1 > 0.5$ is used here to evaluate a question as correct. The first row consists of questions where the individual sub-questions and the whole question were answered correctly. The second row indicates the questions where the final answer was correct despite getting the individual hops wrong, while the third is where the final answer was incorrect despite the individual hops being correct. 10\% of the questions were answered correctly without getting both sub-questions correct. This accounts for over 20\% of the questions that the model got correct, which is considered model failure by \citet{tang-ng-tung-2021-domultihop}, thus indicating that the model indeed follows some form of shortcuts in its multi-hop reasoning process. However, this percentage is much lower than for PLM-based fine-tuned models, which reach close to 50\% \cite{tang-ng-tung-2021-domultihop}.
For 25\% of the questions, the model got both sub-questions correct but was unable to combine them to give the final answer, thus demonstrating difficulties in bridging and integrating separate information during multiple reasoning hops.

\begin{table}[h!]
\begin{center}
\resizebox{1\columnwidth}{!}{%
  \begin{tabular}{lc}
    {\bfseries Type}&{\bfseries Accuracy}\\
    \midrule
    All correct&0.414\\ 
    Correct but sub-questions wrong&0.107\\
    Wrong but both sub-questions correct&0.25\\
  \hline
  \hline
\end{tabular}
}
\caption{Breakdown of the results on running SubQA}
\label{TFF_thing}
\end{center}
\end{table}

\begin{table*}[tb]
\begin{center}
\resizebox{1\textwidth}{!}{%
\begin{tabular}{ll l:l|  ll ll ll ll}
\multicolumn{2}{c}{} & \multicolumn{2}{c}{\bfseries  Overall} & \multicolumn{2}{c}{\bfseries Paragraph Count} & \multicolumn{2}{c}{\bfseries Paragraph Related} & \multicolumn{2}{c}{\bfseries Modified Type} & \multicolumn{2}{c}{\bfseries Second Sub-Q only} \\
\multicolumn{2}{c}{} & EM & F1 & 2 & 4 & Yes & No & Named & Other & No & Yes \\
\hline
    \multirow{2}{*}{\texttt{Llama-2-13B}} & ori & $30.9$ & $45.8$ & $47.6$ & $40.9$ & $47.3$ & $43.6$ & $41.7$ & $46.6$ & $45.9$ & $48.3$ \\
 & adv & $23.6_{-7.2}$ & $33.8_{-12.0}$ & $36.5_{-11.1}$ & $26.1_{-14.7}$ & $32.1_{-15.2}$ & $36.6_{-7.0}$ & $22.2_{-19.5}$ & $35.8_{-10.0}$ & $33.8_{-12.0}$ & $40.6_{-7.7}$\\
 \multirow{2}{*}{\texttt{Mixtral-8x7B-Instruct-v0.1}} & ori & $50.4$ & $68.1$ & $67.9$ & $68.7$ & $67.6$ & $69.0$ & $60.4$ & $69.5$ & $68.2$ & $69.5$\\
& adv & $34.8_{-15.6}$ & $48.4_{-19.7}$ & $50.1_{-17.7}$ & $43.4_{-25.3}$ & $46.2_{-21.1}$ & $52.0_{-17.0}$ & $42.3_{-18.1}$ & $49.5_{-20.0}$ & $48.4_{-19.7}$ & $53.9_{-15.6}$\\
\multirow{2}{*}{\texttt{Llama-2-70b}} & ori & $54.1$ & $67.7$ & $66.1$ & $72.7$ & $65.3$ & $71.6$ & $51.3$ & $70.5$ & $67.7$ & $69.8$\\
& adv & $40.4_{-13.6}$ & $53.2_{-14.5}$ & $53.9_{-12.1}$ & $51.3_{-21.4}$ & $49.7_{-15.6}$ & $58.8_{-12.8}$ & $40.5_{-10.8}$ & $55.4_{-15.1}$ & $53.2_{-14.5}$ & $59.7_{-10.1}$\\
\multirow{2}{*}{\texttt{GPT-3.5}} & ori & $63.4$ & $77.2$ & $76.6$ & $78.5$ & $75.3$ & $80.1$ & $72.9$ & $77.8$ & $77.2$ & $80.5$\\
& adv & $39.9_{-23.4}$ & $52.7_{-24.4}$ & $56.0_{-20.6}$ & $43.1_{-35.4}$ & $51.3_{-23.8}$ & $54.9_{-25.2}$ & $49.7_{-23.3}$ & $53.2_{-24.6}$ & $52.7_{-24.4}$ & $64.3_{-16.1}$\\
\multirow{2}{*}{\texttt{longformer}} & ori & $71.5$ & $82.1$ & $81.0$ & $85.1$ & $80.2$ & $84.6$ & $75.9$ & $83.1$ & $82.1$ & $84.7$ \\
& adv & $51.9_{-19.5}$ & $62.2_{-19.8}$ & $66.1_{-14.8}$ & $50.74_{-34.4}$ & $62.6_{-17.6}$ & $61.6_{-23.3}$ & $62.9_{-13.0}$ & $62.1_{-21.0}$ & $62.2_{-19.9}$ & $61.0_{-23.7}$ \\
\hline
\hline
\end{tabular}
}
    \caption{Results of \texttt{Llama-2-13B}, \texttt{Mixtral-8x7B-Instruct-v0.1}, \texttt{Llama-2-70B}, \texttt{GPT-3.5} and \texttt{longformer} (fine-tuned on the training set) on the original HotpotQA dev set (ori) and our adversarially constructed examples (adv). All the tests for the LLMs are done in the few-shot chain of prompt setting. EM and F1 Performance Scores are reported. F1 scores are further broken down by (left to right): the number of ``fake'' paragraphs; whether ``fake'' paragraphs are related; the type of entity modified, if adversarial paragraphs are unrelated, and if both the adversarial paragraphs are generated from the second sub-question of two different fake sub-question pair.}
\label{result_adversarial}
\end{center}
\end{table*}
\paragraph{III. Reasoning shortcuts in DiRe}
DiRe consists of removing the bridging gold paragraph from the context, with the claim that a model should not be able to answer them under these conditions, and if they are, the examples exhibit a reasoning shortcut exploited by the model. Table \ref{result_DiRe} shows the results of  \texttt{Llama-2-13B} on this. Surprisingly, the model still maintains a decent performance level, confirming that HotpotQA indeed contains several reasoning shortcuts. Seemingly, LLMs---similar to their fine-tuned predecessors---readily exploit such shortcuts despite not being explicitly trained on HotpotQA.

\begin{table}[ht]
\begin{center}
\resizebox{1\columnwidth}{!}{%
\begin{tabular}{lc}
\multicolumn{1}{c}{\bfseries Dataset} & {\bfseries F1 score}\\ 
\hline
\multirow{1}{*}{Original dataset of 4174 examples} & 68.7\\
\multirow{1}{*}{DiRe probe consisting of 5000 examples} & 44.2\\
\hline
\hline
\end{tabular}
}
\caption{\texttt{Llama-2-13b} performance on DiRe when using a normal (non-CoT) prompt and priming with few-shot examples.}
\label{result_DiRe}
\end{center}
\end{table}

\paragraph{IV. Reasoning failures when presented with distracting paragraphs from AddDoc}
Table~\ref{result_adddoc} shows the performance of \texttt{Llama-2-13B}, \texttt{Llama-2-70B} and \texttt{Mixtral-8x7B-Instruct-v0.1}, in few-shot prompt setting,  when attacked with the first 2000 examples of  AddDoc~\cite{jiang2019avoiding}, the most successful method\footnote{ \citet{ding2021reasoning} introduces a attack based on the reasoning chain, but the code/dataset isn't publicly available} to show reasoning weaknesses of models fine-tuned on HotpotQA, by adding crafted paragraphs which are lexically similar to the question.  Apparently, and in stark contrast to fine-tuned models, LLMs performance does not drop on the benchmark, even slightly increasing for some of the evaluated models. 
This finding suggests that the reasoning shortcuts exploited by LLMs are indeed less obvious than simple lexical overlap, thus further motivating the need for a more sophisticated method to evaluate multi-hop reasoning, such as those proposed in this paper. 

\begin{table}[ht]
\begin{center}
\resizebox{1\columnwidth}{!}{%
\begin{tabular}{lcc}
\multicolumn{1}{c}{\bfseries Model} & {\bfseries Original} & {\bfseries AddDoc}\\ 
\hline
\multirow{1}{*}{\texttt{Llama-2-13B}} & 50.3 & 51.7\\
\multirow{1}{*}{\texttt{Mixtral-8x7B-Instruct-v0.1}} & 58.0 & 58.0\\
\multirow{1}{*}{\texttt{Llama-2-70B}} & 53.9 & 54.6\\
\hline
\hline
\end{tabular}
}
\caption{F1 score of \texttt{Llama-2-13b}, \texttt{Llama-2-70b} and \texttt{Mixtral-8x7b-Instruct-v0.1} when attacked with 2000 examples of AddDoc in the few-shot setting.}
\label{result_adddoc}
\end{center}
\end{table}

\subsection{Do LLMs get distracted when faced with seemingly plausible alternatives?}
Table \ref{result_adversarial} shows the results of various open- and closed-source LLMs using our proposed benchmarking method. All models show a significant drop in their F1 scores and their Exact-Match (EM) scores. Importantly, this seems to be a model property rather than an artefact of the prompting technique, as the behaviour persists across different prompting methods (see Appendix~\ref{app:prompting}). 
Furthermore, even GPT-4 exhibits a drop of 14 points in F1 under the strongest adversarial attack setting i.e., when adding four adversarial paragraphs (see Appendix~\ref{app:gpt4}). This is remarkable,  as the benchmark was partially generated with \texttt{GPT-4} in the loop. This highlights the feasibility of our method to evaluate a model using an equally strong model as an adversary, a property that other benchmarks tend to lack \cite{Zellers2018,Zellers2019}.


\subsection{Analysing the effects of different parameters}
Next, we investigate which settings contribute most to the drop in performance.
\paragraph{Count of distractor paragraphs}
As we can modify the number of alternate reasoning chains, and thus generate distractor paragraphs, it is worthwhile investigating whether increasing their number leads to decreased performance. 
Table~\ref{result_adversarial}, ``Paragraph count'' columns, shows the results of the various models in the chain of thought few-shot setting when facing two or four distractor paragraphs, respectively. Indeed, the higher the number of adversarial paragraphs, the more the model struggles, with an additional decrease of about 10 F1 points for every fake reasoning chain on average.


\paragraph{Are the paragraphs related?}
As our method creates fake sub-questions that are used to generate distractor paragraphs, we can modify if the paragraphs to be used in the attack belong to the same fake question pair or not.
If not, the attack will use paragraphs from different pairs but will ensure that if $k$ adversarial paragraphs are being added, $k/2$ are generated from the first sub-question and the other from the second sub-question. This is useful to check if models struggle because of the presence of alternate multi-hop reasoning chains, or if the difference in performance is attributed to distractor paragraphs containing similar but otherwise unrelated information. 

Table \ref{result_adversarial}, columns ``Paragraph Related'' shows the performance of the models in this setting. For \texttt{Llama-2-13B}, \texttt{Mixtral-8x7B-Instruct-v0.1}, and \texttt{Llama-2-70b}, related paragraphs, and therefore complete alternate reasoning chains, cause a larger drop than unrelated distractor paragraphs. Interestingly, \texttt{GPT-3.5} exhibits the opposite behaviour, performing slightly worse when an alternate reasoning chain does not connect the distractor paragraphs. 


\paragraph{Modified type}
Because the main entity of the question can be either part of a Named Entity or not, we can distinguish model performance between these settings.
Table~\ref{result_adversarial}, columns ``Modified Type'', shows the results of this test. Aside from \texttt{Llama-2-13B}, which performs significantly worse on Named Entities, the differences are not statistically significant, indicating that both distractor types seem to be equally difficult. 


\paragraph{Are the paragraphs unrelated and only belong to the 2nd subquestion?}
We have shown that (with the exception of \texttt{GPT-3.5}) examples containing fake paragraphs related by a seemingly alternate reasoning chain are harder for LLMs to process correctly. Similarly, we can investigate if fake paragraphs that are generated purely from the second sub-question add further complexity. Since the paragraph generated from the second sub-question is the only paragraph that contains an entity of the same type as the actual answer, the rationale is to investigate what contributes more to hard multi-hop reasoning: producing seemingly alternate reasoning chains or just adding adversarial paragraphs similar to the paragraph answering the second sub-question. We ensure that the number of adversarial paragraphs, generated using our method, is the same in both settings.


As can be seen in the last column of Table~\ref{result_adversarial}, ``Second Sub-Q only'', all LLMs perform worse when the paragraphs are not generated from the second sub-question only, thus adding further evidence to the hypothesis that examples with seemingly plausible alternate reasoning chains are indeed harder for LLMs to process correctly. Additionally, only the fine-tuned \texttt{longformer} model exhibits the opposite behaviour, suggesting that PLM-based fine-tuned models indeed tend to learn more simple word-matching type heuristics, as generating multiple paragraphs from the second sub-question results in more fake paragraphs that are lexically similar to the question and answer sentence. This adds further evidence that there is a need to reevaluate the weaknesses of LLMs, as insights derived from PLMs do not necessarily carry over.

The second sub-question-only setting is most similar to AddDoc~\cite{jiang2019avoiding} and other existing attacks on HotpotQA. However, unlike for AddDoc, all LLMs still show a drop in performance. This demonstrates the effectiveness of \emph{generating} adversarial paragraphs by changing minute details extracted from the question, surpassing the impact of existing attacks. The paragraphs generated in this manner challenge the LLMs more effectively, highlighting their susceptibility to being ``blinded by nuance''.

\section{Conclusion}
We explored whether LLMs can perform multi-hop reasoning when presented with seemingly plausible yet ultimately incorrect reasoning paths. To do so, we conducted an extensive evaluation to show how LLMs' multi-hop reasoning abilities differ from the previous generation of PLM-based NLP methods relying on fine-tuning. We found that existing adversarial attacks are inadequate to probe the capabilities of LLMs; thus we introduced a simple yet powerful framework based on generating paragraphs that contain seemingly plausible yet wrong alternative reasoning chains, compatible with any benchmark that requires multi-hop reasoning. Our extensive empirical study shows that all evaluated LLMs (including \texttt{GPT-4}) struggle to succeed on the proposed benchmark. The framework facilitates the generation of adversarial paragraphs, enabling the creation of more rigorous tests which could lead to more robust models. Datasets augmented with such adversarial paragraphs could allow the models to move away from learning non-robust features like basic lexical matching and enable improved reasoning capabilities. We release data and code to the wider research community on Github: \url{https://github.com/zawedcvg/Are-Large-Language-Models-Attentive-Readers}.

\section*{Limitations}
The main limitation of the proposed method is that it requires the question to be broken down into its sub-questions. Specifically, we use \citet{tang-ng-tung-2021-domultihop}'s SubQA dataset, but existing question decomposition techniques like \citet{min2019multi} and \citet{perez2020unsupervised} can be used to adapt the framework to all HotpotQA questions or any other dataset that deals with multi-hop reasoning. Furthermore, we use the same algorithm for all types of questions to generate seemingly plausible alternate reasoning paths. However, datasets such as HotpotQA distinguish between different types of multi-hop reasoning, e.g.\ bridge and comparison. Relying on this knowledge, more sophisticated methods to create seemingly plausible alternate reasoning paths could be developed. Although we perform extensive tests to ensure that the current method generates adversarial paragraphs that do not contradict the gold paragraphs, there is no formal guarantee for it.
\bibliography{references, custom}


\appendix
\section{System Prompt for Q/A task}
\begin{tcolorbox}[breakable]
\ttfamily	
\sloppy
\small
You are a helpful, respectful, and honest question-answering assistant. You will be given a context and a question. Answer the question using only the context. You will break the questions into sub-questions. You will then use these sub-questions to get to the final answer. The final answer must have 'Final Answer: ' prepended to it. Thus your output will be in the following format:\\
Sub-question 1: [subquestion 1]\\
Answer: [answer 1]\\
sub-question 2: [subquestion 2]\\
Answer: [answer 2]\\
Sub-question n: [subquestion n]\\
Answer: [answer n]\\
Final Answer: [final answer]\\ The final answer should be limited to 5 words with just the answer and no explanation/information.\\
Here are some past conversations:\\
Context: ......\\
Question: 'Which government position was held by the woman who portrayed Corliss Archer in the film Kiss and Tell'?\\
Sub-question 1: Which woman portrayed Corliss Archer in the film Kiss and Tell?.\\
Answer: Shirley Temple.\\
Sub-question-2: Which government position was held by Shirley Temple?\\
Answer: Chief of Protocol\\
Final Answer: Chief of Protocol.\\
Context : .......\\
Question: What is the name of the fight song of the university whose main campus is in Lawrence, Kansas and whose branch campuses are in the Kansas City metropolitan area?\\
Sub-question 1: Which university has its main campus in Lawrence, Kansas and has branch campuses in Kansas City metropolitan area?\\
Answer: University of Kansas\\
Sub-question 2: What is the name of the fight song of University of Kansas?\\
Answer: Kansas Song\\
Final Answer: Kansas Song\\

\end{tcolorbox}

\section{System Prompt for creating fake paragraphs}
\begin{tcolorbox}[breakable]
\ttfamily	
\sloppy
\small
You are a helpful and respectful fake paragraph generating assistant. You will be given two questions, a few supporting paragraphs, and two words you need to avoid. You will first give a fake answer for the first question. The fake answer should not be the same as any of the two words that need to be avoided. Generate a fake paragraph using the information from the first question and the fake answer generated. The answer and information should not be related to any real-life entity. The paragraphs generated must match the tone of the given two paragraphs. Furthermore, the two paragraphs generated must not contradict any of the information in the supporting paragraphs provided by the user.

Use the fake answer generated for the first question to replace all instances of '[answer]' in the second question. Use the newly generated question and generate a fake answer for it. Ensure that the fake answer generated is not the same as any of the provided words you need to avoid. Similar to the first question, use the fake answer and the question to generate a fake paragraph. You will generate the fake paragraphs as if they were part of a Wikipedia article. You must maintain a neutral and informative tone.

Generate the two paragraphs as separate articles about 75-100 words each. All the answers and paragraphs must be made up of fake names and fake information. The information/names should not reference anyone in real life. Generate exactly one paragraph for each question. Remember to replace all instances of '[answer]' with the answer from the first question and adjust the paragraphs accordingly. However, you must not mention the fact that the details/entities in the paragraphs are fake/imaginary.
\end{tcolorbox}
\section{System prompt for creating fake named entities through GPT-4}
\begin{tcolorbox}[breakable]
\ttfamily	
\sloppy
\small
You are a helpful, respectful and honest fake named entity generator. You will be given upto 20 different entity types along with an example of that type. For each of the entity types, generate another named entity different of the same entity type given the named entity. There are a total of 18 different entity types. The different types and their definitions are as given below:\\
PERSON: People, including fictional\\
NORP: Nationalities or religious or political groups\\
FACILITY: Buildings, airports, highways, bridges, etc.\\
ORGANIZATION: Companies, agencies, institutions, etc.\\
GPE: Countries, cities, states\\
LOCATION: Non-GPE locations, mountain ranges, bodies of water\\
PRODUCT: Vehicles, weapons, foods, etc. (Not services)\\
EVENT: Named hurricanes, battles, wars, sports events, etc.\\
WORK OF ART: Titles of books, songs, etc.\\
LAW: Named documents made into laws\\
LANGUAGE: Any named language\\
DATE: Absolute or relative dates or periods\\
TIME: Times smaller than a day\\
PERCENT: Percentage (including “
MONEY: Monetary values, including unit\\
QUANTITY: Measurements, as of weight or distance\\
ORDINAL: “first”, “second”\\
CARDINAL: Numerals that do not fall under another type\\
For each of the provided examples, you will generate one named entity of the same type.\\

Ensure that your final count of entities is equal to the number of entities in the given prompt. Use indices to help with keeping the count.
\end{tcolorbox}
\section{Dependency type definitions}
Table~\ref{definitions} consists of definitions of the dependency relations used in the attack. All the definitions are based on \citet{de2014universal}
\begin{table}[!htb]
\resizebox{\columnwidth}{!}{%
\begin{tabular}{lp{\columnwidth}}
\hline
Term & Definition \\
\hline
nsubj & nominal subject (nsubj) is a nominal which is the syntactic subject and the proto-agent of a clause. \\
nsubj:pass & A passive nominal subject is a noun phrase which is the syntactic subject of a passive clause. \\
obl & The obl relation is used for a nominal (noun, pronoun, noun phrase) functioning as a non-core (oblique) argument or adjunct. \\
obj & The direct object of a VP is the noun phrase which is the (accusative) object of the verb. \\
acl:relcl & A relative clause (RC) is a clause modifying some head (typically a noun) that is understood to fulfill some grammatical role in the RC. The head is said to be "extracted" from the RC. Most RCs are adnominal, hence the relation acl:relcl. Adverbial RCs attach as advcl:relcl \\
appos & An appositional modifier of a noun is a nominal immediately following the first noun that serves to define, modify, name, or describe that noun. It includes parenthesized examples, as well as defining abbreviations in one of these structures. \\
amod & An adjectival modifier of a nominal is any adjective or adjectival phrase that serves to modify the meaning of the nominal. \\
nmod & The nmod relation is used for nominal dependents of another noun or noun phrase and functionally corresponds to an attribute, or genitive complement. \\
compound & There are noun compounds. \\
flat & The flat relation is used to combine the elements of an expression where none of the immediate components can be identified as the sole head using standard substitution tests \\
\hline
\end{tabular}
}
\caption{Definitions based on \href{https://universaldependencies.org/u/dep/}{Universal Dependencies}}
\label{definitions}
\end{table}

\section{Reproducibility}
\begin{itemize}
       \item The parameters used for fine-tuning the longformer model are
       \begin{itemize}
           \item Batch size: we use batch size of 64 for the longformer model.
           \item Learning Rate: We set learning rate to $3e^{-5}$, as it was reported to work best for the transformer training.
           \item Train Epochs: We train on HotpotQA for 3 training epochs.
           \item Maximal answer length: we set \texttt{max\_answer\_length=30} when obtaining predictions.
       \end{itemize}
       \item All experiments were carried out on a single NVIDIA Titan RTX GPU
\end{itemize}

\section{User study to verify adversarial paragraphs}
\label{subsec:user_study}
To verify that examples don't influence gold labels, a user study\footnote{A copy of the form used for the user-study can be accessed here \url{https://forms.gle/nF8HhoWM48TzcG1K7}} was conducted involving 5 participants, all of whom had at least college level education. Each participant received the same random sample of 49 questions\footnote{There were initially 50 questions, but one question was removed due to an error in creating the form} from the adversarial dataset. It was ensured that the 49 questions were not from the 100 samples that we manually verified. For each question, participants were provided with two sources of information: 
\begin{itemize}
    \item Relevant lines from the gold paragraph.
    \item Relevant lines from adversarial paragraphs intended to distract from the correct answer.
\end{itemize}

The selection of relevant lines followed specific criteria. For the gold paragraphs, we utilized the line numbers identified by the HotpotQA dataset as containing relevant information for the answer. For the adversarial paragraphs, we employed a different approach. During the generation of these paragraphs by GPT-4, plausible answers to sub-questions were crafted. Only sentences containing one of these answers were included, as these would be the sentences that provided information with the potential to mislead the model.

Each question had 4-5 options. One was the correct answer, two were the answers to the sub-questions for the plausible paragraphs, and the rest were titles of the "supporting facts" from HotpotQA if these were not already included in the options. After each question, the user was asked if the two sources of information contained contradicting information. The user was given the following prompt

\begin{tcolorbox}[breakable]
\ttfamily	
\sloppy
\small
Welcome to our user study! In this study, you will be asked to answer 50 questions. Each question will be accompanied by two sources of information. Please read both the question and the provided sources carefully before selecting your answer.

If you encounter any contradictions between the two sources that make it impossible to answer the question accurately, please select "Yes" for "Was there any contradicting information?" Otherwise, select "No". 

Here is an example of contradictory information:
- Source A: "The Kellock-Taschereau Commission was appointed by Adrian Holloway."
- Source B: "The Kellock-Taschereau Commission was appointed by Lyon Mackenzie."

Note: Do not put "Yes" for contradicting information if there is lack of information. You can make assumptions about who the pronouns refer to if the prior isn't mentioned.

Thank you for your participation and careful attention!
\end{tcolorbox}

\begin{table}[ht]
\begin{center}
\resizebox{1\columnwidth}{!}{%
\setlength{\tabcolsep}{30pt} 
\begin{tabular}{lc}
\multicolumn{1}{c}{\bfseries Type} & {\bfseries Accuracy}\\
\hline
\multirow{1}{*}{Average Accuracy} & 70.6\%\\
\multirow{1}{*}{Accuracy-Combined} & 84.6\%\\ 
\multirow{1}{*}{Accuracy-UB} & 95.9\%\\ 
\hline
\hline
\end{tabular}
}
\caption{The three different metrics for accuracy}
\label{result_userstudy}
\end{center}
\end{table}
Table~\ref{result_userstudy} shows three different metrics:
\begin{itemize}
\item \textbf{Average Accuracy}: The percentage of questions answered correctly by the participants.
\item \textbf{Accuracy-Combined}: A question is given 1 point if more than 3 participants answered it correctly, and 0.5 points if exactly 2 participants answered correctly and no incorrect answer got more than 2 votes.
\item \textbf{Accuracy-UB}: Adapted from the HotpotQA paper, this metric checks if any of the users were able to answer a particular question correctly.
\end{itemize}
If a user marks a particular question as containing a contradiction, their answer is marked as incorrect.\\
We use \textbf{Accuracy-Combined} rather than the \textbf{Average Accuracy} as it better deals with the variance in user answers due to misreading certain information. \textbf{Accuracy-Combined} and \textbf{Accuracy-UB} closely follow the results of human-evaluation on HotpotQA~\cite{jiang2019avoiding}. While it is true that our test is simpler than HotpotQA, as the relevant lines are already provided, this test provides strong evidence to suggest that humans aren't affected by the adversarial paragraphs. 

Table~\ref{result_userstudy_confidence} shows the number of questions marked as contradictory with a confidence level of greater than 40\%. The following questions were marked as contradictory with a confidence > 40\% of being contradictory: \href{https://drive.google.com/file/d/1iKozGTG9Q2t48IOHQLRDOfr7KnjrYjhu/view?usp=sharing}{link}. We have gone through the 5 questions marked as contradictory, and just one of them was found to have contradicting information. We attribute the marking of these questions as contradictory to human error and waning attention, which is often present in crowd-sourcing experiments. As per informal user feedback, finding the correct answer was “a bit tricky” at times. Crucially, despite being tricky, they weren’t reported to have contradicting information. Through the accuracy metrics and the count of questions marked as contradictory under different confidence levels, we can conclude with certainty that these distractors do not affect the gold labels and are not an issue for humans.
\begin{table}[ht]
\begin{center}
\resizebox{1\columnwidth}{!}{%
\setlength{\tabcolsep}{20pt} 
\begin{tabular}{lc}
\multicolumn{1}{c}{\bfseries Confidence Level} & {\bfseries Count of Contradictory}\\
\hline
\multirow{1}{*}{40\%} & 5\\
\multirow{1}{*}{60\% or more} & 0\\ 
\hline
\hline
\end{tabular}
}
\caption{The confidence level of a question being contradictory}
\label{result_userstudy_confidence}
\end{center}
\end{table}

\section{Performance of SOTA LLM}
\label{app:gpt4}
To see how well our method generalises to better models, we evaluate \texttt{GPT-4}, the best-performing LLM at the time of writing. \texttt{GPT-4} was tested on 250 examples (due to cost constraints), where fake paragraphs are related by alternate reasoning chains, using Named Entity as the main entity type and alternating between two and four fake paragraphs. Table~\ref{result_gpt4} shows that \texttt{GPT-4} is more resilient to the attack as compared to the other LLMs that were tested. However, it still exhibits a drop of 14 points in F1 under the strongest adversarial attack setting i.e., related paragraphs, four adversarial paragraphs.
\begin{table}[ht]
\begin{center}
\resizebox{1\columnwidth}{!}{%
\begin{tabular}{lccc}

\multicolumn{2}{c}{\bfseries Fake paragraph count} & \bfseries 2 & \bfseries 4\\
\hline
\multirow{2}{*}{\texttt{GPT-4}} & ori & $87.1$ & $91.1$\\
 & adv & $79.9_{-7.2}$ & $77.0_{-14.1}$\\
\hline
\hline
\end{tabular}
}
\caption{F1 scores of the models for 2 and 4 fake paragraphs using \texttt{GPT-4}}
\label{result_gpt4}
\end{center}
\end{table}

\section{Do existing techniques make models more robust?}
\label{app:prompting}
Table~\ref{result_self_consistency} shows the results of running more advanced prompting methods than naive chain-of-thought reasoning, such as instructed chain-of-thought prompting~\cite{shi2023largelanguagemodelseasily} and self-consistency~\cite{wang2023selfconsistency} on the \texttt{Llama-2-13B} and \texttt{Llama-2-70B}. The setting is 2 plausible paragraphs with the modifiable portion as "other". 
While self-consistency leads to a smaller decrease in F1 score under the attack, the gains in robustness (4.2 F1 points) are limited. Instruct prompting on the other hand doesn't provide any relevant improvements. 
This suggests that our findings unveil a behaviour of LLMs that cannot be corrected simply by using more advanced prompting techniques.

\begin{table}[ht]
\begin{center}
\resizebox{1\columnwidth}{!}{%
\begin{tabular}{ccccc}

\multicolumn{2}{l}{\bfseries Prompt Style} & \bfseries CoT & \bfseries COT+Self-consistency & \bfseries COT + Instruct\\
\hline
\multirow{2}{*}{\texttt{Llama-2-13B}} & ori & $39.8$ & $40.1$ & $38.8$\\
 & adv & $20.4_{-19.4}$ & $23.9_{-16.2}$ & $21.5_{-17.3}$\\
\multirow{2}{*}{\texttt{Llama-2-70B}} & ori & $49.4$ & $49.6$ & $49.6$\\
 & adv & $34.4_{-15.5}$ & $36.0_{-13.6}$ & $29.8_{-19.8}$\\
\hline
\hline
\end{tabular}
}
\caption{Effect of self-consistency on F1 score}
\label{result_self_consistency}
\end{center}
\end{table}

\end{document}